\documentclass[11pt,a4paper]{article}

\newif\ifauthordecided
\authordecidedtrue %

\newif\ifarxiv
\arxivtrue %

\newif\ifperfect
\perfectfalse %

\usepackage{times,latexsym}
\usepackage{url}

\usepackage{bm}
\usepackage{bbm} %
\usepackage{amsmath}
\usepackage{amsfonts}
\usepackage{amssymb} %

\usepackage{algorithm,algorithmic}
\usepackage{booktabs}

\usepackage{url}
\usepackage{hyperref}
\usepackage{enumitem}

\newcommand{\red}[1]{\textcolor{red}{#1}}

\usepackage{cleveref}
\crefname{figure}{Figure}{Figures}
\crefname{table}{Table}{Tables}
\crefname{appendix}{Appendix}{Appendices}
\crefname{section}{Section}{Sections}
\crefname{equation}{Eq.}{Eqs.}
\crefname{enumi}{}{} %
\usepackage{pifont}%
\newcommand{\cmark}{\ding{51}}%
\newcommand{\xmark}{\ding{55}}

\newcommand{\zhijing}[1]{ {\color{violet!70}\textit{#1}$_\text{-- Zhijing}$}}

\newcommand{\ourhyp}{{\textit{Bijection Hypothesis}}\xspace}
\usepackage[T1]{fontenc}

\usepackage{amsmath,amsfonts,bm}

\def\eqref#1{equation~\ref{#1}}

\def\1{\bm{1}}

\def\ve{{\bm{e}}}

\def\vp{{\bm{p}}}

\def\vt{{\bm{t}}}

\def\vx{{\bm{x}}}

\def\mT{{\bm{T}}}

\def\mW{{\bm{W}}}
\def\mX{{\bm{X}}}
\def\mY{{\bm{Y}}}

\DeclareMathAlphabet{\mathsfit}{\encodingdefault}{\sfdefault}{m}{sl}
\SetMathAlphabet{\mathsfit}{bold}{\encodingdefault}{\sfdefault}{bx}{n}

\ifarxiv
    \usepackage[acceptedWithA]{tacl2023}
\else 
    \usepackage[]{tacl2023}
\fi

\usepackage{xspace,mfirstuc,tabulary}
\usepackage{times}
\usepackage{latexsym}
\usepackage{amsmath}
\usepackage{graphicx}
\usepackage{subfigure}

\newcommand{\shauli}[1]{\textcolor{blue}{[SR: #1]}}
\newcommand{\yuxin}[1]{\textcolor{orange}{[Yuxin: #1]}}

\newif\iftaclinstructions
\taclinstructionsfalse %
\iftaclinstructions
\renewcommand{\confidential}{}
\renewcommand{\anonsubtext}{  (No author info supplied here, for consistency with
TACL-submission anonymization requirements)}
\newcommand{\instr}
\fi

\iftaclpubformat %

\else

\fi

\title{
\textit{All Roads Lead to Rome?}
\\
Exploring the
Invariance
of 
Transformers' Representations
}

\ifauthordecided
\author{
Yuxin Ren\textsuperscript{\rm 1,}\thanks{\hspace{0.1cm} Done during the research internship with Max Planck Institute \& ETH Zürich.}
\hspace{0.3cm}
Qipeng Guo\textsuperscript{\rm 2} \hspace{0.03cm} 
\hspace{0.2cm}
Zhijing Jin\textsuperscript{\rm 3,4}
\hspace{0.3cm}
Shauli Ravfogel\textsuperscript{\rm 5} \hspace{0.1cm}
\\  %
{\bf 
Mrinmaya Sachan\textsuperscript{\rm 4} \hspace{0.2cm}
Bernhard Schölkopf\textsuperscript{\rm 3,4}
\and 
Ryan Cotterell\textsuperscript{\rm 4} 
}
\\
\textsuperscript{\rm 1}Tsinghua University
{ }
\textsuperscript{\rm 2}Fudan University 
\\
\textsuperscript{\rm 3}Max Planck Institute
for Intelligent Systems, Tübingen, Germany
\\
\textsuperscript{\rm 4}ETH Zürich
{ }
\textsuperscript{\rm 5}Bar-Ilan University
\\
\texttt{jinzhi@ethz.ch} \quad \texttt{ryx20@mails.tsinghua.edu.cn}
\\
}
\fi

\date{}

\begin{document}
\maketitle
\begin{abstract}
Transformer models bring propelling advances in various NLP tasks, thus inducing lots of interpretability research on the learned representations of the models. However, we raise a fundamental question regarding the reliability of the representations. Specifically, we investigate whether transformers learn essentially isomorphic representation spaces, or those that are sensitive to the random seeds in their pretraining process.
In this work, we formulate the \ourhyp, which suggests the use of bijective methods to align different models' representation spaces. We propose a model based on invertible neural networks, BERT-INN, to learn the bijection more effectively than other existing bijective methods such as the canonical correlation analysis (CCA). We show the advantage of BERT-INN both theoretically and through extensive experiments, and apply it to align the reproduced BERT embeddings to draw insights that are meaningful to the interpretability research.\footnote{Our code
\ifarxiv
is at {\url{https://github.com/twinkle0331/BERT-similarity}}.
\else
has been uploaded to the submission system, and will be open-sourced upon acceptance.
\fi
}

\end{abstract}

\section{Introduction}
Transformer models \cite{vaswani2017attention} have emerged as a powerful architecture for natural language processing (NLP) tasks, achieving state-of-the-art results on a wide range of benchmarks \cite[][\textit{inter alia}]{devlin2019bert,gpt3,openai2023gpt4}. However, one issue that has received less attention is the robustness of these models to random initializations. Specifically, it is unclear whether Transformer models learn essentially \emph{isomorphic representation spaces}, or if they learn representations that are sensitive to the random seeds in their pretraining process.

\begin{figure*}[t]
    \centering
    \includegraphics[width=\textwidth]{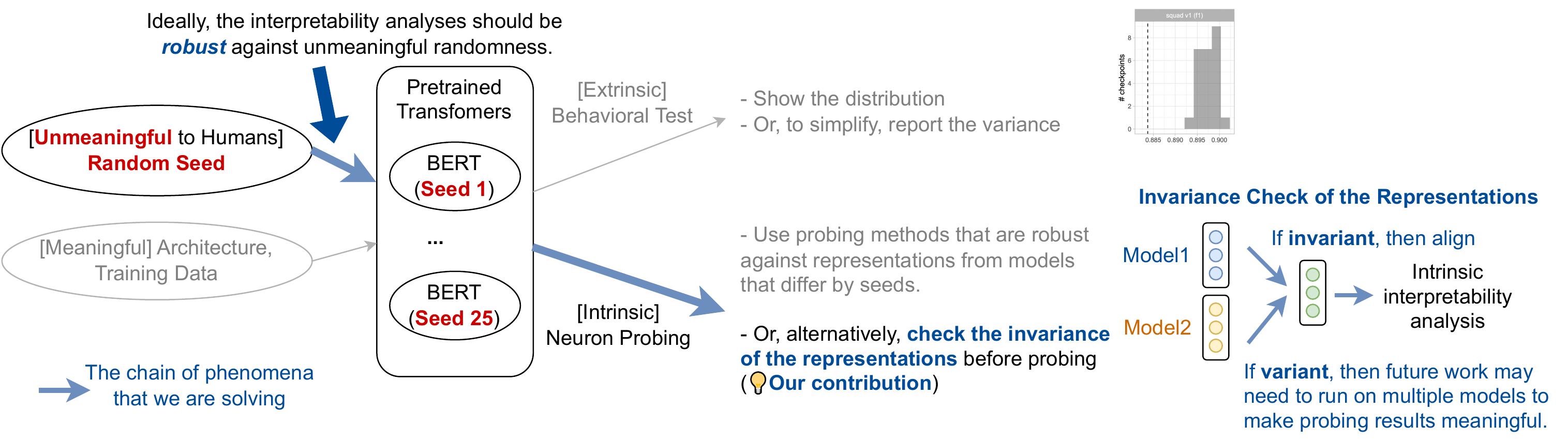}
    \caption{Overview of this work. Specifically, we investigate the similarity of reproduced BERT representations. Given multiple BERT models with different random seeds, we align their representation space according to our \ourhyp, and show clusters of BERT embeddings.
    }
    \label{fig:motivation}
\end{figure*}

Recent studies has shown that small variations in the random seed used for training can result in significant performance differences between models with the same architecture \citep{sellam2021multiberts,dodge2020finetuning}. Understanding the degree to which the learned representation space is sensitive to random initialization is important because it has implications for interpretability research, i.e, understanding \emph{what is encoded} in those models. 
Such efforts can include an analysis of the learned representations of LLMs by probing the information contained within these representations  \citep{tenney2019bert,rogers2020primer}, or by observing the change of neuron activations under intervention  \citep{ravfogel2021counterfactual,meng2022locating,lasri2022probing}. As we illustrate in \cref{fig:motivation}, if a model's learned representation space is highly sensitive to initialization, then it may be necessary to run probing tasks multiple times with different random initializations to obtain more reliable results. Alternatively, techniques to improve the robustness of the learned representation space can be applied to make the models more suitable for probing tasks.

The representation spaces of models trained with different random initializations may be isomorphic due to two main reasons: the properties of the training data or inherent symmetries in the models. For instance, permutation symmetry, which permits the swapping of neurons in the hidden layer without altering the loss function's value, is a typical symmetry observed in neural networks \citep{hecht1990algebraic}. However, prior research has suggested that the permutation symmetry in neural networks may not entirely capture the underlying invariances in their training dynamics \citep{ainsworth2022git}.

This paper introduces the \textbf{\ourhyp}, which posits the existence of a bijection mapping between the representation spaces of different models. To investigate this hypothesis, we propose BERT-INN, an invertible neural network (INN) to align the representation spaces of two BERT models by a set of affine coupling layers. 
We show the advantage of our BERT-INN over four other bijective methods both theoretically and through three different experiments.

Finally, we apply our BERT-INN method to align the embedding spaces of the reproduced BERTs \cite{sellam2021multiberts}, and draw several key findings. Specifically, we identify that (1) the shallower layers are, in general, more invariant than the deeper layers; (2) the interactions of neurons (i.e., attention weights) are more invariant than the representations; and (3) finetuning breaks the variance.

In conclusion, our contributions are as follows:
\begin{enumerate}
    \item We are, to the best of our knowledge, the first to highlight the consistency check of different transformer models' representation spaces.

    \item We pose the \ourhyp, for which we train a BERT-INN model to learn the bijective function mapping and recover the non-linear transformation.
    \item We show the effectiveness our BERT-INN model over four other bijective methods in terms of both theoretical properties and three different empiricial experiments.
    
    \item We align the BERT representations using our BERT-INN method, and reveal several
    insights that have meaningful implications for future work on interpretability research of transformer models.
\end{enumerate}

\section{Background of LLM Analysis}
\label{relatedwork}

\subsection{Large Language Models (LLMs)}
The recent success of NLP is driven by LLMs~ \citep{radford2018improving,devlin2019bert,liu2019roberta,gpt3,ouyang2022instructGPT}.  These models are pre-trained on massive amount of text data, allowing them to learn general language representations that can be fine-tuned for specific NLP tasks. Scaling up the size of language models has been shown to confer a range of benefits, such as improved performance and sample efficiency. LLMs show strong performance in few-shot or even zero-shot tasks and a better generalization of out of distribution tasks.

\subsection{Interpretability of LLMs}

The empirical success of BERT on various NLP domains, encourages researchers to analyse and understand these architectures  \citep{rogers2020primer}.  \citet{kovaleva2019revealing} visualize self-attention modules to probe how tokens in a sequence interact with each other.  \citet{clark2019does} provide in-depth analysis on the self-attention mechanism, showing that some attention heads correspond to linguistic notions of syntax and correference. In addition,  \citet{zhao-etal-2020-quantifying,merchant2020happens} investigate the impacts of downstream task finetuning for BERT representations.  \citet{hewitt-liang-2019-designing,pimentel-etal-2020-pareto} stress that controlling for important factors, e.g., probing classifier's capacity, is crucial to obtain meaningful probing results.

\subsection{The Effect of Random Seeds
}
 \citet{sellam2021multiberts} conduct experiments on pre-trained models with different random seeds for pretraining, and find that the instance-level agreement on downstream tasks varies by the random seeds, and the gap on out-of-distribution tasks is more pronounced.  \citet{dodge2020finetuning} shows that varying only the random seed of fine-tuning process leads to substantial performance gap. The two factors influenced by random seed, training order and weight initalization, contribute equally to the variance of performance.   \citet{zhong-etal-2021-larger} argue that model predictions are
noisy at the instance level. 
On the MNLI dataset, even the same architecture with
different finetuning seeds will lead to different predictions on $\approx 8\%$ of the instances,
due to under-specification  \citep{d2020underspecification}.

\section{The \ourhyp}
\subsection{Representation Spaces of BERTs}

Given two BERT models $\mathcal{M}_1$ and $\mathcal{M}_2$ with identical architecture, but different random initializations, 
the goal is to evaluate their similarity through their representation spaces.

We denote the sets of BERT parameters after the pre-training as 
$\bm{\theta}_1$ and $\bm{\theta}_2$, respectively. Their representation transformation functions are thus $f_{\bm{\theta}_1}$ and $f_{\bm{\theta}_2}$, respectively. 
Since a BERT has 12 transformer layers \cite{vaswani2017attention}, it is a hyperparameter to specify the representation space of which transformer layer we want to analyze. Let us denote the layer of interest as $l$.
Then, the sequence transformation function works as follows.
For any text sequence $\vt$ with $k$ tokens, 
we first get its token-wise embedding $\mT \in \mathbb{R}^{k\times 768}$ at Layer $l$ by
$\mT = \mathrm{Transformer}_{1,\dots,l}(\vt) = \mathrm{FFN}_l(\mathrm{Att}_l(\dots \mathrm{FFN}_1(\mathrm{Att}_1(\mathrm{Emb}(\vt)))))$, where $\mathrm{FFN}_i$ and $\mathrm{Att}_i$ denotes the feedforward layer and the self-attention network at Layer $i$, respectively, and $\mathrm{Emb}$ denotes the embedding layer.
Then we follow the practice in \citet{del2021establishing,wu2019emerging,zhelezniak2019correlations} to do a mean-pooling operation to obtain the overall sentence embedding $\frac{1}{k} \sum_{i=1}^{k} \mT_{i, *}$.

For the two transformation functions $f_{\bm{\theta}_1}$ and $f_{\bm{\theta}_2}$,
we feed a large list of $n$ arbitrary text inputs, $\bm{D} := [\bm{t}_1, \bm{t}_2, \dots, \bm{t}_n]$
to obtain the two sentence representation spaces:
\begin{align}
    \bm{X} &:= [f_{\bm{\theta}_1}(\bm{t}_1),\cdots,f_{\bm{\theta}_1}(\bm{t}_n)] \text{ for }\mathcal{M}_1
    \\
    \bm{Y} &:=[f_{\bm{\theta}_2}(\bm{t}_1),\cdots,f_{\bm{\theta}_2}(\bm{t}_n)]
    \text{ for }\mathcal{M}_2
    ~.
\end{align}
Here, $\mX, \mY \in \mathbb{R}^{n\times d}$, where $d$ is the dimension of these embeddings, or the number of features.

\subsection{Perfect Similarity $=$ Diffeomorphism
}
We propose the \ourhyp, which 
states that two models are perfectly similar if and only if there exists a diffeomorphism in their representation spaces.
\ifperfect
\zhijing{Use 1 sentence to formally define diffeomorphism.}
\fi
The mapping can be discovered through techniques such as linear or non-linear projections. 

Namely, for the two spaces $\mX$ and $\mY$, we learn an invertible function $g$, such that we minimize the distance metric $\delta$ between the transformed $g(\mX)$ and the original $\mY$. 
The similarity of the two spaces can be quantified by the distance metric, and
if $\delta(g(\mX), \mY)$ is zero, then the two spaces are perfectly similar.

\subsection{
Reformulating Existing Similarity Indices into Linear Bijective Functions
}
We briefly introduce existing methods to align the two representation spaces before measuring their similarity. 

\paragraph{Linear Regression.}

An intuitive way to learn a bijective function for aligning two representation spaces is linear regression, which can capture all linear transformations by minimizing the L2 norm. Namely, we learn a matrix $\mW$ which satisfies the following objective:
\begin{equation*}
    \min_{\mW} \delta (\mX\mW,\mY) = \min_{\mW} ||\mX\mW - \mY||_2 
    ~.
\end{equation*}

\paragraph{Canonical Correlation Analysis   (CCA).}

CCA projects the two matrices $\mX$ and $\mY$ into the subspaces which maximize the canonical correlation between their projections. The formulation is as follows: 

\begin{align}
    \rho_i = \max_{\bm{w}_X^i, \bm{w}_Y^i}&  \mathrm{corr}   (  \mX\bm{w}_X^i, \mY\bm{w}_Y^i)\\
 \text{for } &\ ~~ \forall_{j< i}~~ \mX\bm{w}_X^i\perp \mX \bm{w}_X^j
 ~,
 \\
&\ ~~ \forall_{j< i}~~ \mY\bm{w}_Y^i\perp \mY \bm{w}_Y^j .
\end{align}
Here, the objective is to maximize the correlation between all pairs of projected subspaces of $\mX$ and $\mY$, and the constraint is to keep the orthogonality among all the subspaces of each original space. 

Since the dimension of CCA is equal to $d$, by concating the projection vectors into matrices $\mW_{X}$ and $\mW_{Y}$, both $\mX$ and $\mY$ are invertible matrices. The transformation matrix from $\mX$ to $\mY$ is then equivalent to the product of $\mW_X$ and the inverse of $\mW_Y$:
\begin{align}
    \mW_X \mW_Y^{-1}
    ~.
\end{align}

However, since we have to pre-determine the number of canonical correlations $c$, CCA suffers when the guessed $c$ differs from the optimal one.

\paragraph{Noise-Robust Improvements of CCA.}
As CCA might suffer from outliers in the two spaces, there are two types of improvements that are more robust towards outliers or noises in the spaces.

The first variant is singular vector CCA   (SVCCA),
which
performs singular value decomposition (SVD) to extract directions which capture most of the variance of the original space, thus removing the noisy dimensions before running the CCA. 

The other method is projection-weighted CCA   (PWCCA), which computes
a weighted average over the subspaces learned by CCA:
\begin{align}
\rho_\mathrm{PWCCA} &= \frac{\sum_{i=1}^c \alpha_i \rho_i}{\sum_{i=1}^c \alpha_i} & \\
\alpha_i &=\sum_{j} |\langle \bm{p}_i, \bm{x}_j \rangle|
\\
\vp_{i} &= \mX\bm{w}_X^i
~,
\end{align}
where $\vx_j$ is the $j$-{th} feature of the $\mX$ space, namely, the $j$-{th} column. The weight $\alpha_i$ for each correlation measure $\rho_i$ is calculated as how similar the $i$-th subspace projection $\bm{p}_i$ is to all the features $\vx_j$ of the original space $\mX$.

However, SVCCA and PWCCA are still \textbf{\textit{linear}} methods and unable to capture non-linear relationship between $\mX$ and $\mY$.
Further, \citet{csiszarik2021similarity} shows that the low-rank transformation learned by SVD is not as effective as learning a matching under a task loss and the effectiveness of learning a matching stems from the non-linear function in a network.

\subsection{Non-Linear Bijection by INNs
}\label{sec:previous_methods}

All the linear bijective methods introduced previously do not necessarily suit BERTs, as BERT learns non-linear feature transformation through the non-linear activation layers GELU \cite{hendrycks2016gaussian}.
Between two BERT models, there might be a complex, non-linear correspondence between their feature spaces. Hence, we propose the use of INNs to discover bijective mappings between the two BERTs.

INNs learn a neural network-based 
\emph{invertible} transformation from the latent space $\mathcal{U}$ to the observed space $\mathcal{Z}$. 
Using it as our bijective method,
we learn an invertible transformation $g(\cdot)$ from $\mX$ to $\mY$.  Specifically, we optimize the parameters $\bm{\theta}_g$ of the INN function by minimizing the L2 distance $\delta$ between the transformed $g_{\bm{\theta}}   (\mX)$ and $\mY$:
\begin{equation}
    \min_{\bm{\theta}_g}  \delta   (g_{\bm{\theta}_g}   (\mX),\mY)
    ~.
\end{equation}

Intuitively, the invertibility constraint verifies that the mapping between the two representation spaces is \emph{isomorphic}: there is no loss of information when mapping from $\mX$ to $\mY$.

For the implementation of the INN, we draw inspirations from RealNVP \citep{dinh2016density} and adapt it into BERT-INN, which stacks $L$ affine coupling layers to learn the bijective function $g$ to map from the embedding of the text $\vt$ by the first BERT model $f_{\bm{\theta}_1}(\bm{t})$, to that of the second BERT model $f_{\bm{\theta}_2}(\bm{t})$.

In our BERT-INN, each affine coupling layer $\ell$ takes in 
\ifperfect \shauli{I'd just say we use a standard architecture and move the details to the appendix.}
\fi
an input embedding $\ve^{\ell-1}$ of dimension $K$, sets a smaller dimension $k<K$, and produces the output embedding $\ve^{\ell}$ in the following way:
\ifperfect 
is defined by the following equations \red{following the commonly used INN formulation (XXX, 2014)}:
\fi
\begin{align}
\ve^0 &= f_{\bm{\theta}_1}(\bm{t})
\label{eq:inn_init}
\\
\ve^{\ell}_{1:k}& =\ve^{\ell-1}_{1:k} 
\label{eq:inn_keep}
\\
\ve^{\ell}_{k+1:K}&=\ve^{\ell-1}_{k+1:K} \odot \exp   (s^\ell   (\ve^{\ell-1}_{1:k}))+t^\ell   (\ve^{\ell-1}_{1:k})
\label{eq:inn_transform}
~.
\end{align}
In each layer, it keeps the first $k$ dimensions of the input  as in \cref{eq:inn_keep}, and transforms the remaining $K-k$ dimensions by a scaling layer $s^\ell$ and a translation layer $t^\ell$ as in \cref{eq:inn_transform}. And the first layer takes in the embedding 
$\ve^0$ from the representations of the first BERT model as in \cref{eq:inn_init}.
The main spirit of these affine coupling layers is that they form an INN simple enough to invert, but also relies on the remainder of the input vector in a non-trivial manner. 
\ifperfect
\zhijing{Use 1-2 sentences to explain some rationale of why this mapping is meaningful.}
\fi
\ifperfect
\zhijing{Yuxin, can you write the following in terms of equations above? Maybe we can show the "alternation" by $\ve^\ell$ is shuffled in its $K$ dimensions?}
A forward transformation in one affine coupling layer preserves a subset of input vectors, hence an alternating pattern is adopted such that the input vectors that are preserved in one coupling layer are transformed in the next. 
\fi

Finally, the loss objective for the last layer is as follows:
\begin{align}
    \mathcal{L} = || \ve^L - f_{\bm{\theta}_2}(\bm{t})||_2
    ~,
\end{align}
where we optimize for the L2 distance $||\cdot||_2$ between the last layer's embedding $\ve^L$ and the second BERT's representations $f_{\bm{\theta}_2}(\bm{t})$.

\section{Comparing the Bijective Methods
}
In this section, we show the effectiveness of our proposed INN method against all other bijective methods in terms of the theoretical properties, and empirical performance across synthetic data and real NLP datasets.

\subsection{Experimental Setup}\label{sec:setup}
\paragraph{Reproduced BERT Models} 
We conduct the experiments using the 25 reproduced BERT models
released by \citet{sellam2021multiberts}. The BERT models vary only by random seeds, but adopt the same 
training procedure as in the original BERT model \citep{devlin2019bert}.

\paragraph{Datasets}  
To get the BERT representations, we run the 25 trained BERT models in the inference mode, and plug in a variety of datasets and tasks. Specifically, we choose three diverse tasks from the commonly used GLUE benchmark \cite{wang2019glue}: 
the natural language inference dataset
MNLI \citep{williams2018broad},
the sentiment classification dataset
SST-2  \citep{socher2013recursive},
and the paraphrase detection dataset MRPC  \citep{dolan2005automatically}.
These datasets also cover a variety of domains from
news, social media, and movie reviews, which are the common domains for many NLP datasets.
The dataset statistics are in \cref{tab:data}.
\begin{table}[t]
    \centering \small
    \setlength\tabcolsep{4pt}
    \begin{tabular}{llccl}
    \toprule
    Dataset & Task & Train & Test & Domain \\ \midrule
    MNLI & Inference & 393K & 20K
    & Mixed \\
    SST-2 & Sentiment & 67K & 1.8K & Movie reviews \\
    MRPC & Paraphrase & 3.7K & 1.7K & News \\
    \bottomrule
    \end{tabular}
    \caption{Dataset statistics from the GLUE benchmark \cite{wang2019glue}.}
    \label{tab:data}
\end{table}

\paragraph{Baselines}
We compare against four commonly used existing bijective methods introduced in \cref{sec:previous_methods}, namely linear regression, CCA, SVCCA, and PWCCA. Note that PWCCA sometimes generates results same as or similar to CCA, because it learns the mapping using the same principles as the CCA, and then computes a weighted average over the coefficients, which could generate similar results to CCA under some conditions.

\paragraph{Evaluation Setup}
For each method $g$, we evaluate their alignment performance by measuring the distance between two spaces: the ground-truth target space $\bm{Y}$, versus the learned space $g(\bm{X})$ transformed from the original $\bm{X}$ with the alignment method $g(\cdot)$.
We adopt the a widely used distance function, Euclidean distance, also known as the L2 distance, to measure the distance between the two spaces, namely $\delta(g(\bm{X}), \bm{Y}) := ||\bm{Y} - g(\bm{X})||_2$. Since our goal is to align the two spaces, the smaller the L2 distance, the better, with the best value being zero. For SVCCA, since it reduces the dimension of the original spaces, we compute the L2 norms between the dimension-reduced matrices. To show the effectiveness of our INN method, we randomly select eight model pairs to report the average of their pair-wise L2 distances using each of the bijective methods. And after we verify the effectiveness of our INN method, we apply it on all 25 models to draw overall insights on the similarity of the model representations.

\subsection{Advantage of INN in Theoretical Properties}\label{sec:theoretical_properties}
\begin{table}[ht]
    \setlength\tabcolsep{0.5pt}
    \centering \small
    \resizebox{\columnwidth}{!}{
    \begin{tabular}{lccc}
    \toprule
         & Transformation Type & Non-Lin. & Noise-Robust \\ \midrule
    Lin. Reg. & {\fontsize{7}{4}\selectfont $||Q_Y^\text{T}X||_\text{F}^2/||X||_\text{F}^2$} & \xmark & \xmark \\
    CCA & {\fontsize{7}{4}\selectfont $||Q_Y^\text{T}Q_X||_\text{F}^2/d$}  & \xmark & \xmark \\
    SVCCA & {\fontsize{7}{4}\selectfont $||(U_Y T_Y)^\text{T}U_X T_X||_\text{F}^2/\text{min}(||T_X||^2_\text{F},||T_Y||^2_\text{F})$} & \xmark & \cmark \\
    PWCCA & {\fontsize{7}{4}\selectfont $\sum_{i=1}^{d} \alpha_i \rho_i/||\mathbf{\alpha}||_1$
} 
& \xmark & \cmark \\ \hline
    INN (Ours) & Any invertible function & \cmark & \cmark \\
    \bottomrule
    \end{tabular}
    }
    \caption{Theoretical properties of all the alignment methods. We focus on properties such as whether the method can handle non-linear mapping between the two spaces (``Non-Lin.''), and whether the method is robust against noise in the transformation (``Allow Noise'').}
    \label{tab:theoretical_comparison}
\end{table}

We compare the theoretical properties of all the alignment methods in \cref{tab:theoretical_comparison}. 
\ifperfect \shauli{I'd consider removing this table. I'd also consider removing the following paragraph. Both CCA and regression are linear, so the comparison doesn't make that much sense.}
\fi
As a result of the nature of the transformation types, we can see 
that all previous methods can only handle linear transformations, whereas our INN can model non-linear, invertible mappings. Moreover, linear regression and CCA are sensitive towards noise in the embedding space; SVCCA and PWCCA improves upon CCA by filtering out the noise; and our model is also robust against noises in that the training objective of INN is to minimize the overall loss across all data points.
\ifperfect, and is robust against outliers. \shauli{I don't understand this argument. INN uses a variant of the $L_2$ loss which is by definition sensitive to outliers.}\zhijing{I agree. Yuxin, do we have a better way to highlight INN's advantage?}
\fi

\subsection{Advantage of INN on Synthetic Data
}

We also show empirical evidence of the effectiveness of our method.
For our empirical experiment,
we want to understand whether INNs are capable of {identifying} the correspondence between the two representation spaces. We cannot do that with raw BERT representations, since we do not know how invariant the BERT representations are, and also we do not know its form of bijection, if such invariance exists.
Therefore we generate synthetic data where the correspondence is known, and we measure the ability of INNs to identify it.

\begin{table}[ht]
    \centering
    \small
    \begin{tabular}{lcc}
    \toprule
    & L2 Distance $\delta(g(\bm{X}), \bm{Y})$ ($\downarrow$) \\ \midrule
    Linear Regression & 1.0882{\tiny $\pm$0.3867}  \\
    CCA & 0.3060{\tiny $\pm$0.1948} 
    \\
    SVCCA & 0.1396{\tiny $\pm$0.1246} \\
    PWCCA & 0.3060{\tiny $\pm$0.1948} \\
    INN (Ours) & \textbf{0.0059}{\tiny $\pm$0.0037} \\
    \bottomrule
    \end{tabular}
    \caption{
    Empirical performance of all the alignment methods to recover a
    non-linear, invertible transformation. For L2 distance ($\downarrow$), the smaller, the better, and the best possible L2 distance is zero, which means the learned mapping recovers the ground-truth transformation between the two spaces.
    }
    \label{tab:syn}
\end{table}
\begin{table*}[ht]
    \centering
    \small
    \begin{tabular}{lcc@{\extracolsep{7pt}}cc@{\extracolsep{7pt}}ccccccc}
    \toprule
         & \multicolumn{2}{c}{MNLI}  & \multicolumn{2}{c}{SST2} & \multicolumn{2}{c}{MRPC} \\ \cline{2-3} \cline{4-5} \cline{6-7}
         & Train & Test & Train & Test & Train & Test \\
         \midrule
    CCA & 0.1484{\tiny $\pm$0.0763} & 0.1534{\tiny $\pm$0.0802} & 0.3064{\tiny $\pm$0.1791} & 0.2535{\tiny $\pm$0.1784} & 0.1455{\tiny $\pm$ 0.0663} & 0.1502{\tiny $\pm$0.0668} \\
    SVCCA & 0.1216{\tiny $\pm$0.1011} & 0.1180{\tiny $\pm$0.0989} & 0.2399{\tiny $\pm$0.2322} & 0.1997{\tiny $\pm$0.2375} & 0.1033{\tiny $\pm$0.1136} & 0.0852{\tiny $\pm$0.0730} \\
    PWCCA & 0.0546{\tiny $\pm$0.0441} & 0.0529{\tiny $\pm$0.0418} & 0.1095{\tiny $\pm$0.1214} & 0.0877{\tiny $\pm$0.1080} & 0.0424{\tiny $\pm$0.0344} & 0.0391{\tiny $\pm$0.0288} \\
    Lin. Reg. & 0.1003{\tiny $\pm$0.0799} & 0.1024{\tiny $\pm$0.0822} & 0.0884{\tiny $\pm$0.0533} & 0.0896{\tiny $\pm$0.0623} &  0.0618{\tiny $\pm$0.0392} & 0.0641{\tiny $\pm$0.0424} \\
    INN (Ours) & \textbf{0.0520}{\tiny $\pm$0.0388} & \textbf{0.0461}{\tiny $\pm$0.0336} & \textbf{0.0563}{\tiny $\pm$0.0315} & \textbf{0.0523}{\tiny $\pm$0.0401} & \textbf{0.0299}{\tiny $\pm$0.0165} & \textbf{0.0310}{\tiny $\pm$0.0180} \\
    \midrule
    \textit{Sanity Check:}\\
    Non-Bij. NN & 0.0360{\tiny $\pm$0.0256}  & 0.0371{\tiny $\pm$0.0272}  & 0.0479{\tiny $\pm$0.0279}  & 0.0449{\tiny $\pm$0.0352}  & 0.0247{\tiny $\pm$0.0145} & 0.0263{\tiny $\pm$0.0165}  \\
    \bottomrule
    \end{tabular}
    \caption{Alignment performance on MNLI, SST2, and MRPC datasets using CCA, SVCCA, PWCCA, linear regression (Lin. Reg.), and our INN method. We report the L2 norm $\delta$ ($\downarrow$) after aligning the sentence space on the training set (just for a reference) and test set (as the main result). The training and test samples were randomly selected from the respective datasets, with a sample size of 1600. For sanity check, we also report the performance of a non-bijective neural network (Non-Bij. NN), which should lead to the smallest L2 norm because it has less constraint than INN, although at the expense of violating the \ourhyp. 
    \ifperfect
    \shauli{Reminder: need to show \textbf{relative} reconstruction errors across the paper.} \yuxin{Since the  rerunning all experiment takes too much time, here we only compute the mean MSE loss among the 12 layers. The inputs keep the same for all methods, so the comparsion is fair. For the boxplot figure in \cref{fig:pretrain_hidden}, we show the relative reconstruction errors.} \shauli{Not sure I understand. Why does it matter that the input is the same? two models can differ in mean feature magnitude across all layers.} \yuxin{We keep the input to be the same, we only change the models. Here we study the invariance of representations across models. The MSE error of the corresponding layer is divided by the norm(magnitude) of the features.}
    \fi
    }
    \label{tab:performance}
\end{table*}

Specifically, we make the first space $\bm{X}$ a real BERT representation space generated by \citet{sellam2021multiberts}. For the second subspace $\bm{Y}$, we generate it by applying a randomly initialized INN $f$ to the first space $\bm{X}$, namely $\bm{Y} = f(\bm{X})$ as the ground truth.

Then, without any knowledge of the transformation $f$, we apply all methods to obtain their corresponding mapping function $g$ given the two spaces $\bm{X}$ and $\bm{Y}$.
For each method, we evaluate their alignment performance by measuring the distance between the learned $g(\bm{Y})$ and ground-truth $\bm{Y}$ using L2 distance, namely $\delta(g(\bm{Y}), \bm{X}) := ||g(\bm{Y}) - \bm{X}||_2$.
Note that since the ground truth mapping $f$ gives an L2 distance of zero, when interpreting the results, the lower L2 distance, the better an alignment method is.

We show the performance of our method compared with all the baselines in \cref{tab:syn}.
All baselines (linear regression, CCA, SVCCA, and PWCCA) struggle to learn a good mapping between the two spaces, with an L2 distance always larger than 0.13, which is quite large.
In contrast, our INN method gives a very small L2 distance, 0.0059, which almost recovers the ground-truth transformation.
The results of this experiment echoes with the theoretical properties analyzed previously in \cref{sec:theoretical_properties}.

\subsection{Advantage of INN on Real Data's 
Representation Similarity
}
We now move from synthetic data to real NLP datasets. As introduce in \cref{sec:setup}, we use three diverse datasets: MNLI \citep{williams2018broad},
SST-2  \citep{socher2013recursive},
and MRPC  \citep{dolan2005automatically}.

We report the alignment results in \cref{tab:performance}. For each method, we use the splits in the original data, learn the mapping using the training set, (if applicable,) tune the hyperparameters on the validation set, and finally report the performance on the test set. For some additional references, we also report the training set performance (to check how well the method fits the data), in addition to the test set performance (as the final alignment performance).

Among all the bijective methods in \cref{tab:performance}, our INN method leads to the smallest L2 distance on all datasets by a clear margin. We can also see that the vanilla CCA is usually the weakest, followed by SVCCA. PWCCA and linear regression learns the bijective mapping relatively well, but still falling behind our proposed INN method.

\ifperfect
Additionally, since each BERT has 12 layers, we visualize the L2 norm across all layers 
in \cref{fig:sim_comparsion}. The visualization provide a more fine-grained view into the layer-wise alignment quality behind the overall numbers reported in \cref{tab:performance}. And consistent to \cref{tab:performance}, INN can effectively align the neurons across layers with a much lower L2 distance.
\fi

\subsection{Advantage of INN on Real Data's Behavioral Similarity
}

The results presented thus far showcase the effective mapping capability of INNs between the representation spaces of two models. However, it remains unclear how applying this mapping would impact the actual \emph{behavior} of the models. If an accurate bijection between the representation spaces of two models is established, it should be possible to \emph{intervene} during the forward pass of one model by substituting its representation with the corresponding representation from the other model, without causing significant loss. In this section, we conduct a behavioral analysis to investigate this hypothesis.

We design the following ``layer injection'' experiment: 
For each pair of BERTs $\mathcal{M}_1$ and $\mathcal{M}_2$ finetuned on a downstream classification task (e.g., MNLI, SST2, or MRPC), we denote the embedding spaces of the last layer before the classification as $\bm{X}$ for model $\mathcal{M}_1$, and $\bm{Y}$ for model $\mathcal{M}_2$.
We learn a transformation function $g$ to map $\bm{X}$ to $\bm{Y}$.
, we generate the transformed embedding $g(\bm{X})$ and inject it into the second model $\mathcal{M}_2$.
With this direct replacement, we keep all the other weights in $\mathcal{M}_2$ unchanged, and run it in the inference mode to get the performance on all three datasets.
Our experiment design carries a similar spirit to the existing work in computer vision on convolutional neural networks \citep{csiszarik2021similarity}.

\begin{table}[ht]
    \centering
    \small
    \setlength\tabcolsep{2.5pt}
    \begin{tabular}{lcccc}
    \toprule
         & MNLI  & SST2 & MRPC \\
         \midrule
    Random Emb. & 33.12{\tiny $\pm$14.63} & 53.97{\tiny $\pm$26.83} & 42.49{\tiny $\pm$23.89} \\
    Random Label & 33.55{\tiny $\pm$0.43} & 50.44{\tiny $\pm$1.20} & 52.9{\tiny $\pm$2.51} \\
    Lin. Reg. & 37.15{\tiny $\pm$24.26} &  54.14{\tiny $\pm$31.27} & 45.93{\tiny $\pm$28.48} \\
   CCA & 30.37{\tiny $\pm$9.30} & 45.46{\tiny $\pm$12.36} & 51.72{\tiny $\pm$18.60} \\
    PWCCA & 39.70{\tiny $\pm$4.57} & 50.24{\tiny $\pm$0.90} & 50.00{\tiny $\pm$18.38}
    \\
    Non-Bij. NN & 
    29.61{\tiny $\pm$15.37} & 64.25{\tiny $\pm$19.78} & 53.58{\tiny $\pm$25.76} 
    \\
    INN (Ours) & \textbf{75.75}{\tiny $\pm$8.97} & \textbf{74.01}{\tiny $\pm$15.61} & \textbf{86.34}{\tiny $\pm$1.07} \\ 
    \bottomrule
    \end{tabular}
    \caption{Behavioral check (by accuracy on the validation set) of various bijective mapping methods across three datasets, MNLI, SST2, and MRPC. Apart from linear regression (Lin. Reg.), CCA, PWCCA, non-bijective neural network (Non-Bij. NN), and our INN method, we also include two random baselines, one using random labels, and the other using random embeddings (Random Emb.), where we apply an untrained, randomly initialized classifier on top of the pretrained BERT encoder.}
    \label{tab:func_check}
\end{table}

As we can see in \cref{tab:func_check}, our model generates the best performance across all datasets, which is quite impressive compared with the close-to-random performance of many other methods.
Moreover, our performance is even better than that of the non-bijective neural networks, which might indicate that the non-bijective transformation is not the proper way to go. It might distort some features in the embedding space, making it incompatible with the following layers of the injected BERT. Previous work also shows a discrepancy between representational similarity and behavior similarity.

\section{Findings: Variance and Invariance across BERTs}
In this section, we utilize the INN alignment method on the 25 reproduced BERTs to address the following question:
    \textit{How similar are these BERTs (which only differ by random seeds)?
    }

This question holds significance for the NLP community because both answers matter for future work directions: If these BERTs are similar, then all the work built on one single BERT will generalize to other BERTs. And if these BERTs differ, then follow-up work on BERT should also test their findings on the other BERTs varied by random seeds, to make sure the findings are robust.

To answer this overall question, we ask three subquestions from different perspectives:
(1) Across the 12 representation layers, which ones are similar, and which ones are different?
(2) How does the way how hidden states interact with each other (i.e., attention weights) vary?
(3) How does finetuning enlarge or reduce the differences?

\subsection{Deep vs. Shallow Layers}
\begin{figure}[h]
    \centering
    \includegraphics[width=\columnwidth]{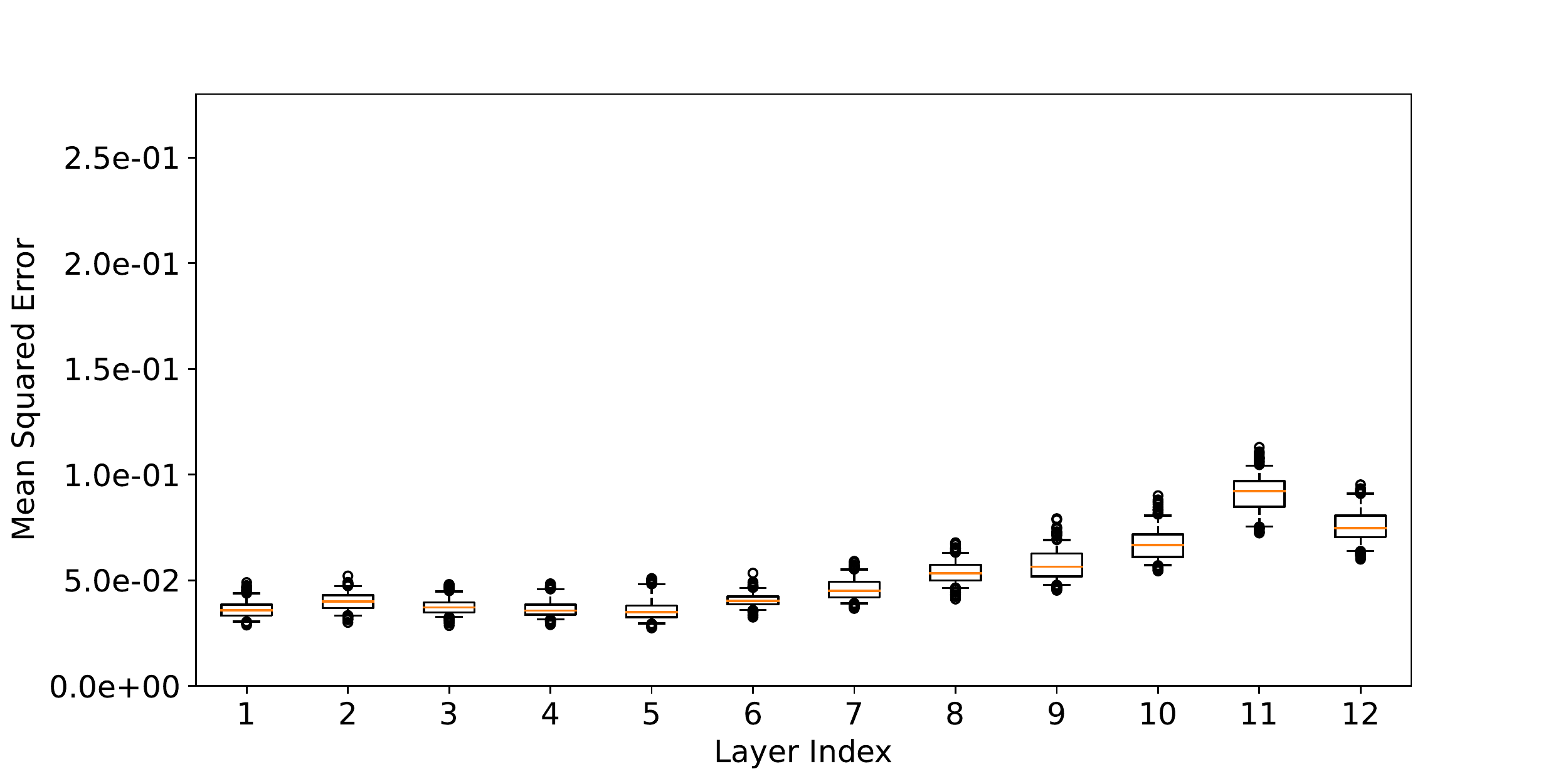}
    \caption{
    Similarity of the \textit{hidden states} learned in the 12 layers of BERTs. For each layer, we align the 25 BERTs and plot the distribution of their L2 distances after alignment. 
    The shallow layers (Layer 1 -- 6) are similar, with a small L2 distance of around 0.02, whereas the deep layers (Layer 7 -- 12) start to be more different, all the way to a large L2 distance of over 0.06.
    }
    \label{fig:pretrain_hidden}
\end{figure}

We plot in \cref{fig:pretrain_hidden} the similarity of the representation spaces across the 12 layers of the BERTs. We show here the experimental results on the largest dataset in our previous experiments, MNLI, and show in appendix the experiments on other datasets.

As we can see, the shallow layers (Layer 1 -- 6) are relatively similar, all having an L2 distance close to 0.02, which is very small. However, the deeper layers (Layer 7 -- 12) start to see an increase in inconsistency, with a larger L2 norm which rises from 0.03 to over 0.06, tripling the distance of shallow layers.

\begin{figure}[t]
    \centering
    \includegraphics[width=\columnwidth]{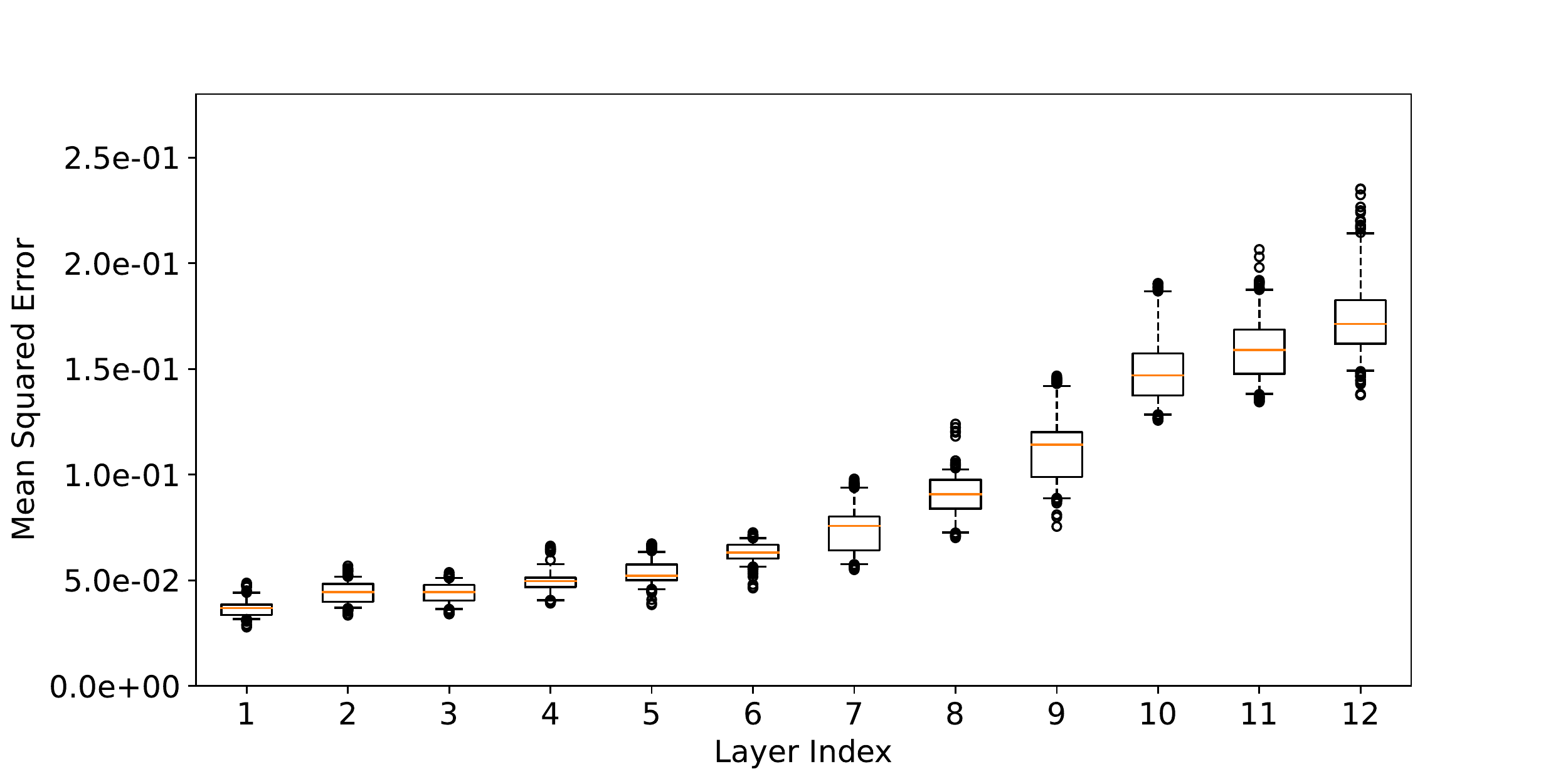}
    \caption{
    Similarity of the 12 layers of BERTs. We can see that the transformation in different layers are more similar for layers next to each other, and differ a lot when the layers are far away from each other.
    }
    \label{fig:layer_similarity}
\end{figure}
\begin{figure}[t]
    \centering
    \includegraphics[width=\columnwidth]{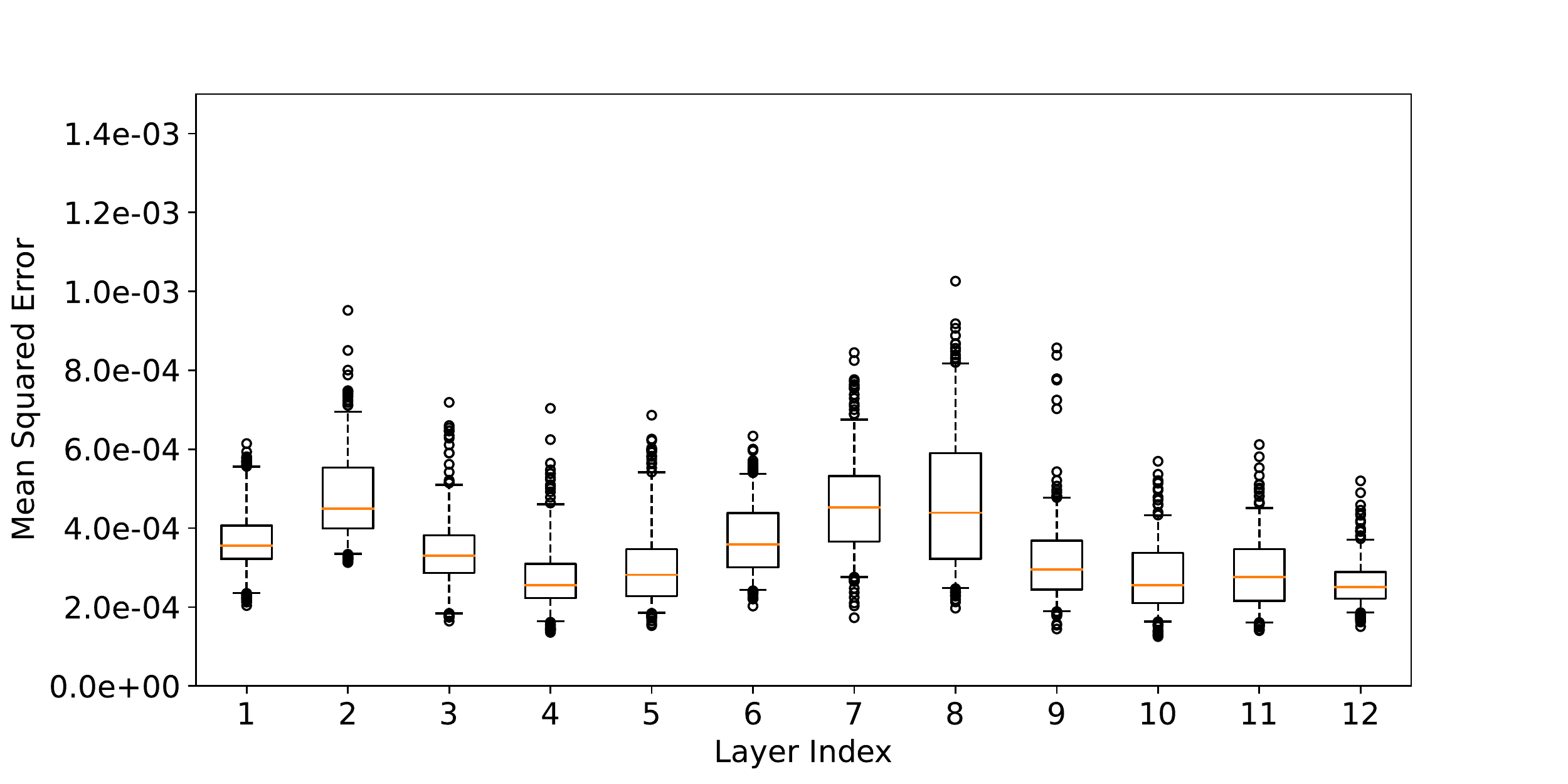}
    \caption{
    Similarity of the \textit{attention weights} in the 12 layers of BERTs. For each layer, we align the 25 BERTs and plot the distribution of their L2 distances after alignment.
    }
    \label{fig:pretrain_attention}
\end{figure}
One potential explanation for the observed results is that the shallower layers of the model tend to learn more low-level features that are shared across various texts, whereas the deeper layers are capable of capturing more complex semantic features that can have multiple plausible interpretations. To support this claim, we present the results of our INN method, which measure the similarity across all 12 layers, in Figure \ref{fig:layer_similarity}. 

There are three notable findings from our analysis:
(1) Two areas of high similarity, represented by whiteness, are present in the plot, suggesting that the representations of the shallow layers are similar among themselves, as are the representations of the deep layers. (2) The last 
shows similarity 
to the 7th layer, since it needs both low-level features and high-level features in order to predict the word correctly. 
(3)  The difference in the representation space changes gradually across the layers, without any abrupt changes.

\subsection{Representation vs. Interaction}

The transformer architecture of BERT learns not only the hidden representations, but also the way how the hidden states should \textit{interact}, which are expressed by the weight matrices in the attention layers.
Hence, we inspect the similarity of the attention weights in \cref{fig:pretrain_attention}. Surprisingly, the attention weights are far more invariant than the representations, with L2 distances lower by two orders of magnitude, ranging from 0.0002 to 0.0004.
This indicates that the way the hidden states interact tend to be very similar across all BERTs.

\subsection{Before vs. After Finetuning}

\begin{figure}[t]
    \centering
    \subfigure[Similarity of the \textit{hidden states} \textit{after} fintuning. We can see that the shallow layers stay invariant, whereas the deep layers vary even more than those before finetuning in \cref{fig:pretrain_hidden}.]{
    \begin{minipage}[t]{0.8\columnwidth}
    \centering
    \includegraphics[width=3in]{figures/box_fin_MNLI_val_hid_INN_scale.pdf}
    \label{fig:finetune_hidden}
    \end{minipage}
    }
    \\
    \subfigure[Similarity of the \textit{attention weights} \textit{after} fintuning. We can see that the invariance is mostly kept across all layers.]{
    \begin{minipage}[t]{0.8\columnwidth}
    \centering
    \includegraphics[width=3in]
    {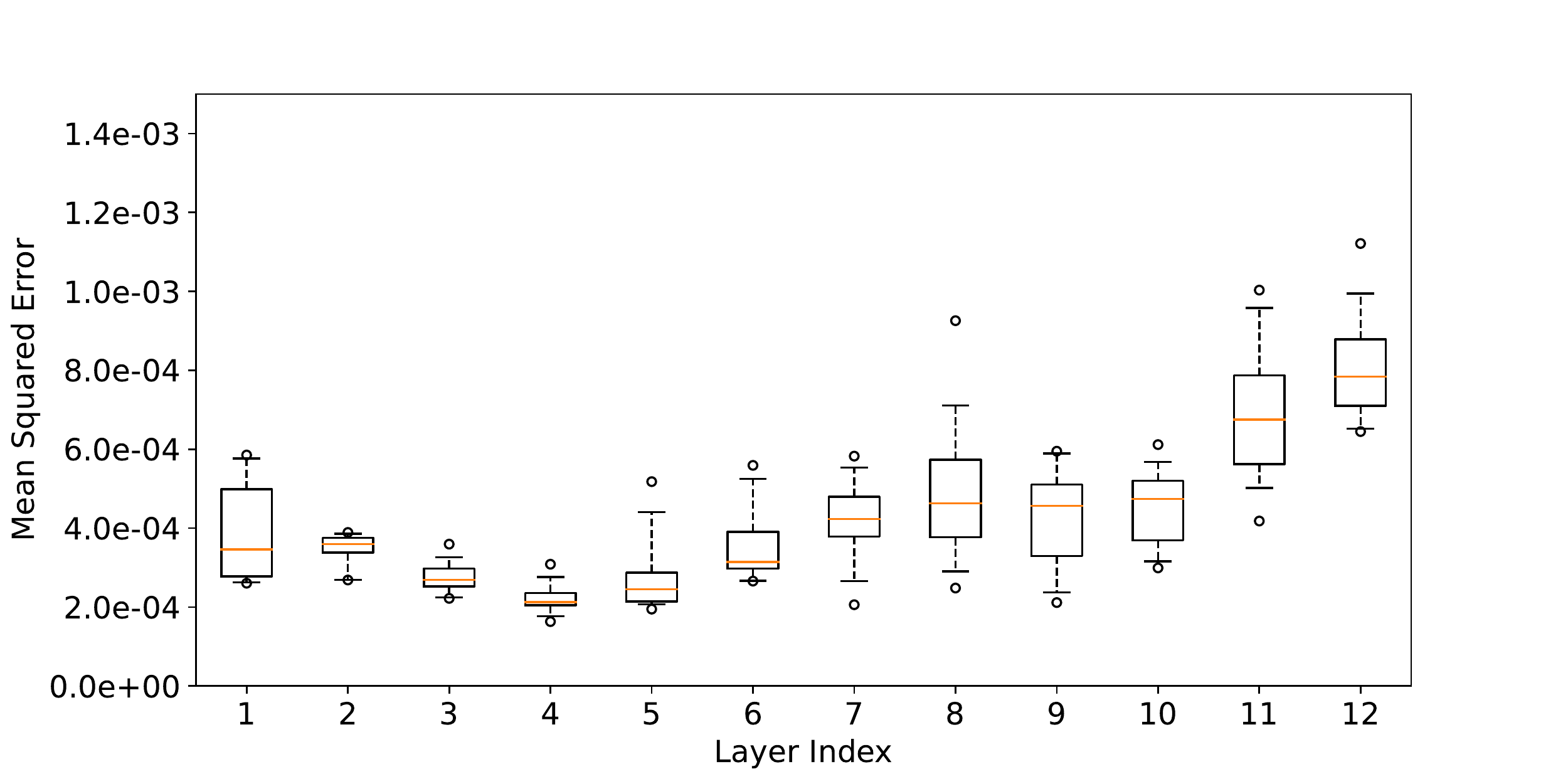}
    \label{fig:finetune_attention}
    \end{minipage}
    }
    \caption{   
    Similarity across the 12 layers of BERTs after finetuning. The deep representations tend to vary more, whereas the shallow representations and attention mechanisms tend to stay similar.
    }
    \label{fig:finetune}
\end{figure}
Finally, we inspect how finetuning affects the similarity among BERTs.
In \cref{fig:finetune_hidden}, we can see that, although the shallow layers (1 -- 6) remain similar after finetuning, the deep layers (7 -- 12) vary drastically, reaching a high L2 distance of up to 0.12, which doubles its original maximum L2 distance of 0.06 in \cref{fig:pretrain_hidden}.
On the contrary, the way how the hidden states interact stays more invariant, as we can see that the attention weights stay similar across the BERTs, with L2 distances still mostly around 0.0002 to 0.0004 in \cref{fig:finetune_attention}.

\section{Conclusion}
In this work, we explored the question of how invariant are transformer models that differ by random seeds. We proposed the \ourhyp, for which we investigated a set of bijective methods to align different BERT models. Through theoretical and empirical analyses, we identified that our proposed INN is the most effective bijective method to align BERT embeddings. Applying the INN method, we identify that shallow layers of BERTs are more consistent, whereas deeper layers encode more variance, and finetuning brings a substantial variance to the models, which imply that the finetuning process very possibly leads the model to different alternative local minima. Our work brings a unique insight for future work interpreting the internal representations of transformer models.

\section*{Ethical Considerations}
This work aims to measure the similarity of the learned representation space of the commonly used NLP model, BERT. This is a step towards safer model deployment, and more transparency of the black-box LLMs.
As far as we are concerned, there seem to be no obvious ethical concerns for this work. All datasets used in this work are commonly used benchmark datasets in the community, and there is no user data involved.

\ifarxiv
\section*{Author Contributions}
This project originated from a discussion between Zhijing and Qipeng in 2020. Inspired by Bernhard's causal representation learning \cite{
scholkopf2021towards} and promising results in computer vision, they did some preliminary exploration into running ICA on BERT embeddings as a first step of causal representation learning for NLP. 
When Yuxin joins MPI and ETH as an intern with Zhijing, he discovers that ICA shows some results that could vary a lot by the random seed, which inspires this work on inspecting the what random seeds bring to the BERT models.
Yuxin then did extensive analysis into the model representations, and, then, with the insights of Ryan, we together formulate this \ourhyp. Yuxin completed all the analyses over 300 model pairs and many iterations of experiments. Shauli added insights when we change from CCA methods to INN, as well as many additional experiments such as exploring the behavioral similarity. Mrinmaya and Bernhard co-supervised the project. And all of the coauthors contributed to the writing.

\section*{Acknowledgment}

This material is based in part upon works supported by the German Federal Ministry of Education and Research   (BMBF): Tübingen AI Center, FKZ: 01IS18039B; by the Machine Learning Cluster of Excellence, EXC number 2064/1 – Project number 390727645; by the John Templeton Foundation (grant \#61156); by a Responsible AI grant by the Haslerstiftung; and an ETH Grant
(ETH-19 21-1).
Zhijing Jin is supported by PhD fellowships from the Future of Life Institute and Open Philanthropy.
\fi

\bibliography{anthology,custom,sec/refs_ai_safety,sec/refs_causality}
\bibliographystyle{acl_natbib}
\end{document}